\documentclass[letterpaper, 10 pt, conference]{ieeeconf}  

\IEEEoverridecommandlockouts                              

\usepackage{amsmath,amsfonts}
\usepackage{algorithmic}
\usepackage{algorithm}
\usepackage{array}
\usepackage{textcomp}
\usepackage{stfloats}
\usepackage{url}
\usepackage{verbatim}
\usepackage{graphicx}
\usepackage{cite}
\usepackage{soul}
\usepackage{xcolor}
\usepackage{booktabs} 
\usepackage{multirow}
\usepackage{amsmath}

\usepackage{float}
\usepackage{caption}

\usepackage{hyperref}

\def\figref#1{Fig.~\ref{#1}}

\def\eqref#1{Eq.~(\ref{#1})}

\title{\LARGE \bf
VLM-Empowered Multi-Mode System \\for Efficient and Safe Planetary Navigation
}

\author{Sinuo Cheng$^{1}$, Ruyi Zhou$^{1}$, Wenhao Feng$^{1}$, Huaiguang Yang$^{1}$, Haibo Gao$^{1}$, Zongquan Deng$^{1}$, Liang Ding*$^{1}$
\thanks{This work was supported in part by the National Natural Science Foundation of China under Grant 52425502, and in part by the National Key R\&D Program of China under Grant 2022YFB4702300, 
and in part by the National Natural Science Foundation of China, Basic Research Program for Young Students (Doctoral Students) under Grant 523B2039, and in part by the Fundamental Research Funds for the Central University under Grant HIT.OCEF.2023042, and in part by the Self-Developed Projects of the National Key Laboratory of Aerospace Mechanism under Grant 2024ASM-ZY03, and in part by the National Natural Science Foundation of China under the Basic Science Center Program under Grant T2388101.(\textit{Corresponding author: Liang Ding.})}
\thanks{$^{1}$These authors are with the State Key Laboratory of Robotics and System, Harbin Institute of Technology, Harbin 150080, Heilongjiang, China. (e-mail:{\tt\small sinuo\_cheng@126.com}, {\tt\small zhouryhit@gmail.com},{\tt\small fengwh\_hit@163.com}, {\tt\small yanghuaiguang\_hit@163.com},{\tt\small gaohaibo@hit.edu.cn}, {\tt\small dengzq@hit.edu.cn}, {\tt\small liangding@hit.edu.cn})
        }%
}

\begin{document}

\maketitle
\thispagestyle{empty}
\pagestyle{empty}

\begin{abstract}

The increasingly complex and diverse planetary exploration environment requires more adaptable and flexible rover navigation strategy. In this study, we propose a VLM-empowered multi-mode system to achieve efficient while safe autonomous navigation for planetary rovers. Vision-Language Model (VLM) is used to parse scene information by image inputs to achieve a human-level understanding of terrain complexity. Based on the complexity classification, the system switches to the most suitable navigation mode, composing of perception, mapping and planning modules designed for different terrain types, to traverse the terrain ahead before reaching the next waypoint. By integrating the local navigation system with a map server and a global waypoint generation module, the rover is equipped to handle long-distance navigation tasks in complex scenarios. The navigation system is evaluated in various simulation environments. Compared to the single-mode conservative navigation method, our multi-mode system is able to bootstrap the time and energy efficiency in a long-distance traversal with varied type of obstacles, enhancing efficiency by 79.5\%, while maintaining its avoidance capabilities against terrain hazards to guarantee rover safety. More system information is shown at https://chengsn1234.github.io/multi-mode-planetary-navigation/.
\end{abstract}

\section{INTRODUCTION}

The exploration activities of planetary rovers at different investigation locations on planetary surface rely on their abilities to navigate across the environment. Researches have shown that challenges such as uneven topography, wheel slippage, and soil deformation can significantly affect rover mobility safety and energy or time efficiency. Assuring safe mobility while employing more efficient navigation strategies can accelerate the exploration process and enhance the efficiency of scientific investigations by maximizing traversable areas and reducing time spent on hazard avoidance.
 
Contemporary planetary rover navigation has evolved from heavy reliance on ground-commanded instructions towards a high degree of embodied autonomy. China’s Zhurong Mars rover has successfully demonstrated its short-range autonomous navigation capability at Utopia Planitia \cite{ding2022surface}. NASA’s Curiosity rover has incorporated vision-based hazard detection and autonomous path planning to enhance its ability to navigate through complex terrain, with different planning and control strategies used to enhance its adaptability\cite{rankin2021mars}. The Perseverance rover, equipped with the latest navigation system upgraded AutoNav, significantly improved its autonomy, enabling it to traverse and explore its surroundings more effectively, and exceeded the previous Mars single-day distance record (219 m, set by Opportunity in 2005) \cite{verma2023autonomous}. These systems implement advanced navigation strategies that allow rovers to perceive, analyze, and make informed decisions either before or during traversal.

As planetary surface on the Moon or Mars is complex in geometry and varied in properties, multi-mode navigation strategies are increasingly adopted to improve both efficiency and safety. For example, for flat and solid terrain, simple navigation strategy like moving directly point-by-point without perception is enough, while for terrains scattered with multi-type obstacles, it requires a more comprehensive system with additional perception capabilities and adaptive avoidance strategies to move safely. Locomotion modes can also be divided into different levels to improve the navigation efficiency. For instance, on safe surface, the rover can be commanded in higher speed to facilitate faster traversal, while in more hazardous and challenging landscapes, rover needs to move more cautiously (at a conservative speed) in order to support more complex, time-consuming calculations and navigation strategies to ensure safety. 

Despite these advancements, current multi-mode navigation systems often rely heavily on human intervention for mode assignment. For example, the Curiosity rover employs a dual-mode system where it can either execute specific computer commands designed by engineers or autonomously map danger zones and plan the safest route based on the complexity of the terrain\cite{rankin2021mars}. However, the analysis of terrain complexity in this system still depends largely on engineers on Earth, and the design of modes is relatively rudimentary, lacking sophisticated considerations for seamless transitions between modes. Moreover, the discontinuity in mode switching leads to inefficiencies and potential risks, showing the need for an automated system that can adapt to changing terrains real-time.

Recent advancements in Vision-Language Models (VLMs) have significantly enhanced their ability to parse complex scene information by integrating visual and linguistic cues. These models can process images, identify objects, understand spatial relationships, generate contextual descriptions with a level of comprehension comparable to humans, and even reason about unseen scenarios. With the ability to achieve human-level understanding of visual scenes, VLMs are becoming powerful tools which can replace the role of human in terrain complexity analysis to make accurate decision-making in navigation strategies in scientific exploration applications based on visual perception.

\begin{figure*}[!b]
   \vspace{-1.0em}
   \centering
   \includegraphics[scale=1.0]{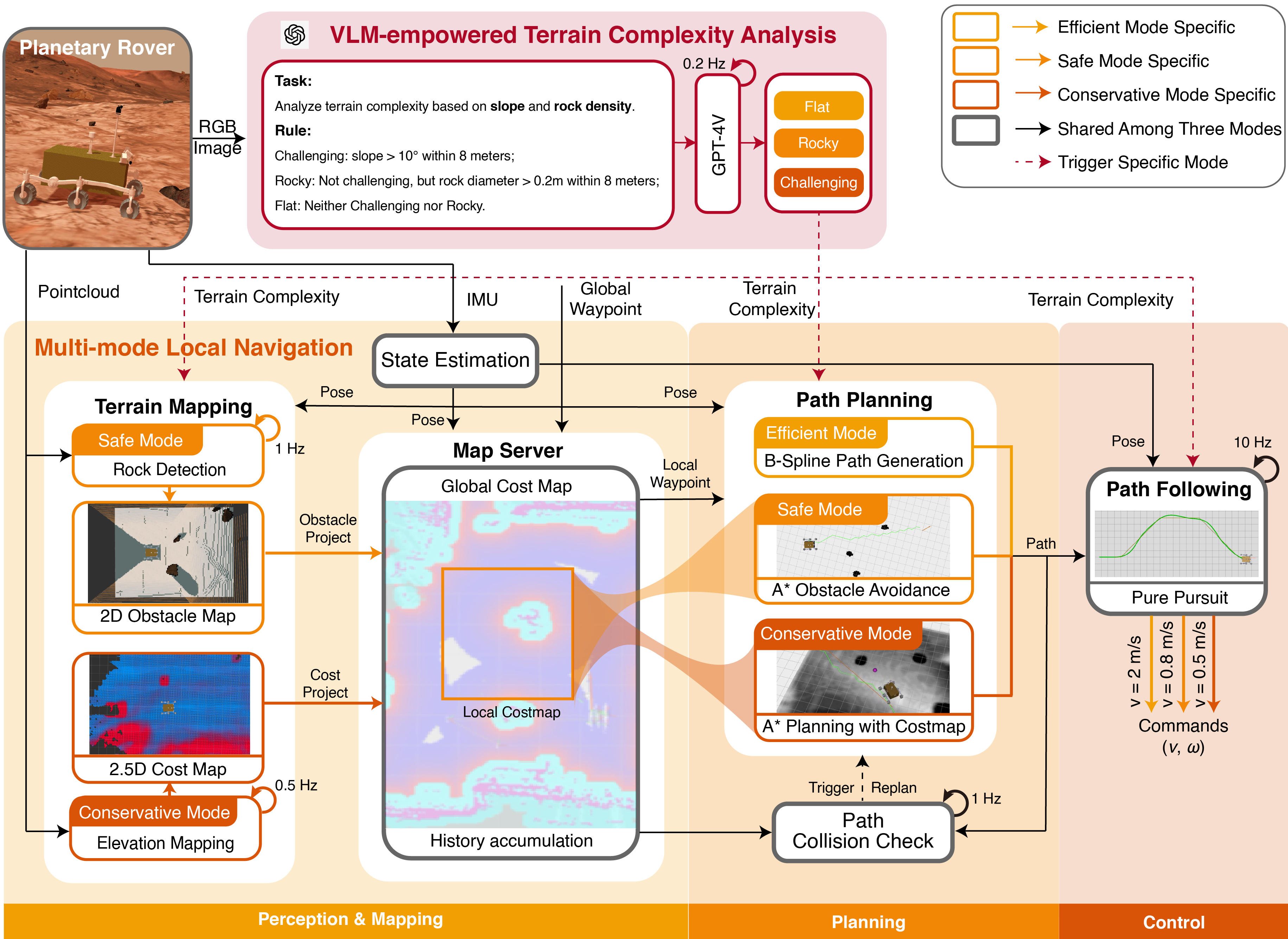}
   \caption{Overview of the multi-mode local navigation framework for planetary rovers.}
   \label{main_figure}
\end{figure*}

Integrating all these ideas, this study develops a multi-mode navigation framework (\figref{main_figure}) combined with a speed-tiered approach and VLM-empowered terrain complexity analysis to further enhance the autonomy and efficiency of planetary rover navigation. By tailoring navigation methods to specific terrain types and dynamically adjusting movement speeds, the proposed system aims to reduce traversal time and improve navigation efficiency on complex terrain, while assuring safe navigation through hazards in diverse terrain conditions.

The main contributions of our work are as follows:
\begin{itemize}
  \item \textbf{A multi-mode navigation framework for planetary rovers} by switching to different complexity of navigation strategies according to the terrain conditions without manual intervention during navigation. 
  \item \textbf{Vision-Language Model empowered terrain complexity analysis}, providing a more human-comprehensible understanding of terrain compared to traditional geometric methods.
  \item \textbf{Experimental analysis and quantitative evaluation} in simulation, demonstrating the superior performance of our system in efficiency while ensuring safe traversal across challenging terrain.
\end{itemize}

\section{Related Work}
\subsection{Multi-mode Navigation Systems for Planetary Rovers}

In real missions, the Mars Exploration Rovers (MERs), Spirit and Opportunity, introduced multi-mode navigation with manual commands, directed driving and autonomous path selection\cite{biesiadecki2006mars}. Operators could select terrain estimation, path selection and visual odometry methods, resulting in eight different modes, providing obstacle avoidance but required human supervision in complex environments\cite{leger2005mars}. Curiosity extended multi-mode autonomy with three visual odometry modes, four path selection strategies and two wheel control methods\cite{rankin2021mars}, which could dynamically switch between slip control and full odometry, adjust paths using terrain data, and adopt terrain-responsive wheel speed control (TRCTL), but still required frequent stops in hazardous areas. Perseverance achieved real-time multi-mode autonomy, enabling continuous adaptation to the terrain\cite{verma2023autonomous}. It introduced five driving modes, including fully autonomous AutoNav, semi-autonomous Guarded Mode and directed driving with or without hazard checks. Using real-time hazard detection and neural-network based navigation, it completed 88.7\% of its 17.7km journey autonomously, setting a record for continuous driving. Unlike Curiosity, which relied on intermittent re-evaluations, Perseverance continuously adjusted its path for smoother, faster exploration.

Researchers have also conducted a number of studies on efficient multi-mode navigation methods for planetary rovers. In general, navigation modes are determined based on mission complexity and terrain characteristics, and so far can be categorized into Blind Driving, Obstacle Avoidance, and Terrain Mapping for different scenarios\cite{azkarate2020gnc}\cite{weclewski2022sample}. Blind Driving mode uses rapid movement to enable the rover to reach a specified destination quickly. Obstacle Avoidance mode uses visual or radar sensing to detect and navigate around small obstacles at relatively high speed. Terrain Mapping mode focuses on building a local terrain map, taking into account factors such as terrain traversability, slip rates and other critical information to determine and follow a safe path. Some navigation systems are being enhanced on this basis, such as using different field-of-view and resolution settings for local mapping in different terrain\cite{marc2018autonomous}. These researches indicate that multi-mode navigation systems are now increasingly being adopted in planetary exploration to optimize the balance between navigation efficiency and safety.

\subsection{Terrain Traversal Complexity Analysis}

Previous planetary missions have relied on generic methods for local map construction and terrain traversability analysis. The Spirit and Opportunity rovers used the GESTALT system, which processed 3D pointcloud to identify geometric hazards such as rocks, ditches, and cliffs, based on criteria like step, tilt, and roughness\cite{maimone2007overview}. Curiosity enhanced terrain perception by integrating high-resolution image from the Mars Reconnaissance Orbiter (MRO) with NavCam-generated terrain meshes. Its onboard software performed continuous safety checks, hazard avoidance, and traversability analysis to maintain safe navigation \cite{rankin2021mars}. Building on these approaches, the Perseverance rover employs advanced terrain perception methods, including visual odometry (VO) with stereo cameras for position estimation, Approximate Clearance Evaluation (ACE) to assess wheel-ground contact, and slip prediction based on terrain slope and wheel rotation \cite{verma2023autonomous}. These methods enable Perseverance to navigate autonomously and safely on the challenging Martian terrain. 

Research on terrain classification methods has been carried out for traversability assessment and path planning through geometric and semantic approaches. Seraji et al. from JPL proposed the Fuzzy Traversability Index (FTI) to classify terrain complexity into three levels—simple, moderate, and complex—based on terrain slope, roughness, and hardness using a fuzzy logic framework, though the membership functions for key parameters were empirically designed \cite{seraji2000fuzzy}. C. Brooks et al. from MIT classified terrain traversability into six levels using a dimensionless traction coefficient as the evaluation metric \cite{brooks2009learning}. Based on terrain classification results, they employed online learning techniques to infer traversability and generate terrain maps \cite{ono2016data}. G. Hedrick et al. developed an efficient path planning algorithm for Martian day driving, categorizing terrain into five traversability classes, including benign, rough, sandy, no-AutoNav, and untraversable, each consisting of two distinct terrain types \cite{hedrick2020terrain}.

In spite of diversity and widespread application of terrain classification methods, these approaches often come with high computational costs due to mapping and processing requirements. Moreover, the methods struggle to mimic human understanding effectively, especially in ambiguous regions where manual intervention still remain necessary. These limitations suggest the need for more adaptive and efficient terrain understanding frameworks that can better align with human-like decision making in complex environments.

\section{Multi-Mode Local Navigation System}
\subsection{System Overview}

The local navigation system consists of a VLM terrain classifier and three separate navigation methods, coping with different types of terrain. The VLM terrain classifier determines the traversal complexity by evaluating the slope and rock distribution based on RGB images, and the terrain complexity is classified into three categories--flat, rocky, and challenging. For easy flat terrain, the rover follows a shortest path using a relatively high speed, maximizing efficiency. For rocky terrain, the rover employs rock detection for local obstacle avoidance. For more challenging terrain, a conservative strategy using 2.5D mapping and cost planning is performed, assuring safe traversal. By dynamically assessing terrain complexity and adapting to navigation strategy real-time, the rover secures both efficiency and safety across diverse environments. 

\subsection{VLM-empowered Terrain Complexity Classifier}

Planetary environments are highly diverse and complex, including various terrain features and potential hazards. Based on the type of hazards, we categorize the terrain into three distinct classes. The friendliest one is flat terrain with minimal obstacles and elevations, offering the least traversal difficulty. The next one is rocky terrain with minor elevation variations but includes obstacles such as scattered rocks, untraversable bumps or pits that require careful obstacle avoidance. Lastly, the challenging terrain is a combination of geometric hazards and significant elevation changes, including steep slopes and hills, posing a challenge for conservative navigation.

In our method, GPT-4V\cite{yang2023dawn}, a Vision-Language Model (VLM) developed by OpenAI, is employed to conduct the classification process. RGB images are served as inputs, and the module is prompted to analyze two terrain features: rock distribution and slope, based on the hazard type we mentioned above. An example prompt is illustrated as \figref{main_figure}, where the terrain is classified into three distinct categories. To enhance the interpretability of the results, the classifier is further required to score the complexity of rock distribution and slope variations from 0-1 separately, reflecting VLM's understanding of terrain. In the end, a strictly JSON format response is required, and result is published as a ROS topic for other modules. Compared to traditional geometric methods, the prompt-based approach is more intuitive and understandable to humans, and can be flexibly adapted to different vehicles and missions without extensive parameter tuning. 

\subsection{Efficient Navigation Mode for Flat Terrain}

For flat terrain, the goal is to maximize navigation efficiency by allowing the rover to reach local waypoints quickly. This mode eliminates onboard perception and complex planning, reducing computational load and achieving high-speed operation. It generates smooth paths and uses a simple algorithm for efficient movement toward the target. 

Based on vehicle's current position and navigation waypoint, a smooth navigation path is generated using the B-Spline method. A pure pursuit algorithm\cite{wallace1985first} is then employed for path following with the control rate of 10Hz, which is simple in principle and satisfies real-time requirements of navigation. In this mode, a relatively high navigation speed is set, maximizing efficiency. Additionally, dynamic look-ahead distance is adopted to allow the rover to smoothly follow the path.

\subsection{Safe Navigation Mode for Rocky Terrain}

For rocky terrain, the rover must efficiently reach the waypoint while avoiding collisions. This mode relies on onboard sensors to detect obstacles and construct a local obstacle map for real-time avoidance, traversing safely while maintaining efficiency. However, the navigation speed is reduced for onboard processing and cautious obstacle avoidance.

\subsubsection{Obstacle map construction} The system detects obstacles in the surrounding area using pointcloud data. Based on the height differences, rocks and pits are extracted and marked as untraversable obstacles. These obstacles are stored in a local obstacle map represented in the occupancy map format, with an update rate of 1 Hz.

\subsubsection{A* planning and obstacle avoidance}  An A* local planner is active during navigation, planning a path to the waypoint based on the obstacle map. The path planning process is triggered if a new waypoint is received, as well as if current path intersects with obstacles in the local obstacle map, to maintain a collision-free path. Lower navigation speed is set compared to the efficient mode in order to carry out real-time obstacle avoidance.

\subsection{Conservative Navigation Mode for Challenging Terrain}

For terrains with significant elevations and obstacles, a safe traversal is of most importance. This mode relies on precise terrain evaluation and path planning, maximizing safety in complex environments. However, higher computational load results in a slower movement to allow real-time processing and safe path-following, reducing navigation efficiency. 

\subsubsection{Terrain mapping and cost calculation} The mapping module is developed based on Elevation Mapping Cupy \cite{miki2022elevation, erni2023mem} , which uses GPU acceleration and can easily adapt to different sensor inputs. To evaluate the traversal cost of the terrain, an elevation gridmap is constructed using depth pointcloud, and the traversal cost of each grid is calculated based on the geometric feature, with the update rate of 0.5 Hz.

\subsubsection{A* planning with costmap} Based on the costmap, we implement an A* search to plan a path with the lowest traversal cost to the local waypoint. A cost threshold is set, so that replan was triggered only if collision is detected or the cost is above tolerance. A conservative navigation speed is set to assure stable path-following, avoiding deviations from the path that may lead to danger.

\section{Closed-loop System Integration and Implementation Details}

The closed-loop navigation system consists of a map server, a waypoint initialization module, pose estimation, and the local navigation part afore-introduced. The map server maintains a global costmap during the whole navigation process; the waypoint initialization module initializes a set of local waypoints before the navigation; local navigation implements the navigation methods developed in the previous section, continuously updating and requesting local map from the map server, receiving rover state from pose estimation, performing navigation methods, guiding the rover along the waypoints to the destination.

\subsection{Map Server and Waypoint Initialization}

The map server is designed to centrally manage the global costmap and waypoints, making sure that all algorithms have consistent access to data through a unified interface. A key feature of the map server is its ability to continuously update the costmap from the active navigation mode for history accumulation, allowing higher-mode to override lower-mode map data. Additionally, by utilizing ROS services, it standardizes interactions across all navigation algorithms, simplifying integration, scalability, and future extensions.

The map server maintains the global costmap in the form of OccupancyGrid, with values from 0 to 100 representing traversal costs, and -1 for unknown areas. The global costmap is initialized before the mission, and remains active throughout the whole navigation process. During the navigation, the server continuously performs map update based on received local map, and provides local map and waypoint to the planning module if requested. Additionally, path collision checks are implemented and performed at the rate of 1 Hz, and the system sends a replan signal if collision is detected. As the navigation proceeds, global costmap is continuously updated and accumulated, while giving a priority for map data provided by higher modes.

Waypoint generation is performed before the navigation either manually or automatically. Human operators can give a set of local waypoints, or waypoints can be generated by computing traversal cost based on satellite elevation map, and through a cost-minimizing search from start to goal. The waypoint set is then stored in the server and given to the planning module in order, guaranteeing that the rover is navigating through the waypoints sequentially.  

\subsection{Local Navigation Implementation}

For our mission, the rover is 3.3m×3.2m×1.85m in size, designed to operate on the surface of Mars. The VLM prompt and parameters of navigation algorithms are adjusted in order to cope with the mission and vehicle dimensions. Pose estimation is carried out using IMU data, providing rover state information to navigation modules for mapping, planning and control. During the navigation, the local navigation system activates the corresponding mode based on VLM classification result,  which is updated at the frequency of 0.2 Hz and broadcasted to other navigation modules. The system continuously exchanges data with the map server for real-time updates. For flat terrain, no map update to the map server is performed, and local waypoint is received to generate an efficient path. For rocky terrain, an 20m×20m local obstacle map with the resolution of 0.5m is constructed, and then projected into the global costmap of the map server. Binarized local map along with the local waypoint is received for local planning. For complex terrain, an 20m×20m local costmap with the resolution of 0.1m is updated to the map server, and local costmap is received for planning. Based on the mission and mode characteristics, the path-following speeds for three navigation modes are set to 2.0 m/s, 0.8 m/s, and 0.5 m/s respectively.

\section{Simulation Environment}
\subsection{Simulation Platform}
The simulation experiments are carried out on MarsSim \cite{zhou2022marssim}, which is developed based on Gazebo. It provides both physical and visual realistic simulations for planetary rovers, supporting testing navigation algorithms in the Martian environment. Based on the platform, we are able to create a comprehensive set of terrain elements that can impact the navigation, including geometrically defined obstacles such as rocks of various sizes and randomly distributed cracks, large-scale elevation changes like rolling hills and steep slopes, and small-scale terrain irregularities such as surface wrinkles and uneven textures.

\subsection{Terrain Design}

\begin{figure}[!tbp]
\vspace {6pt}
\centering

\includegraphics[scale=1.0]{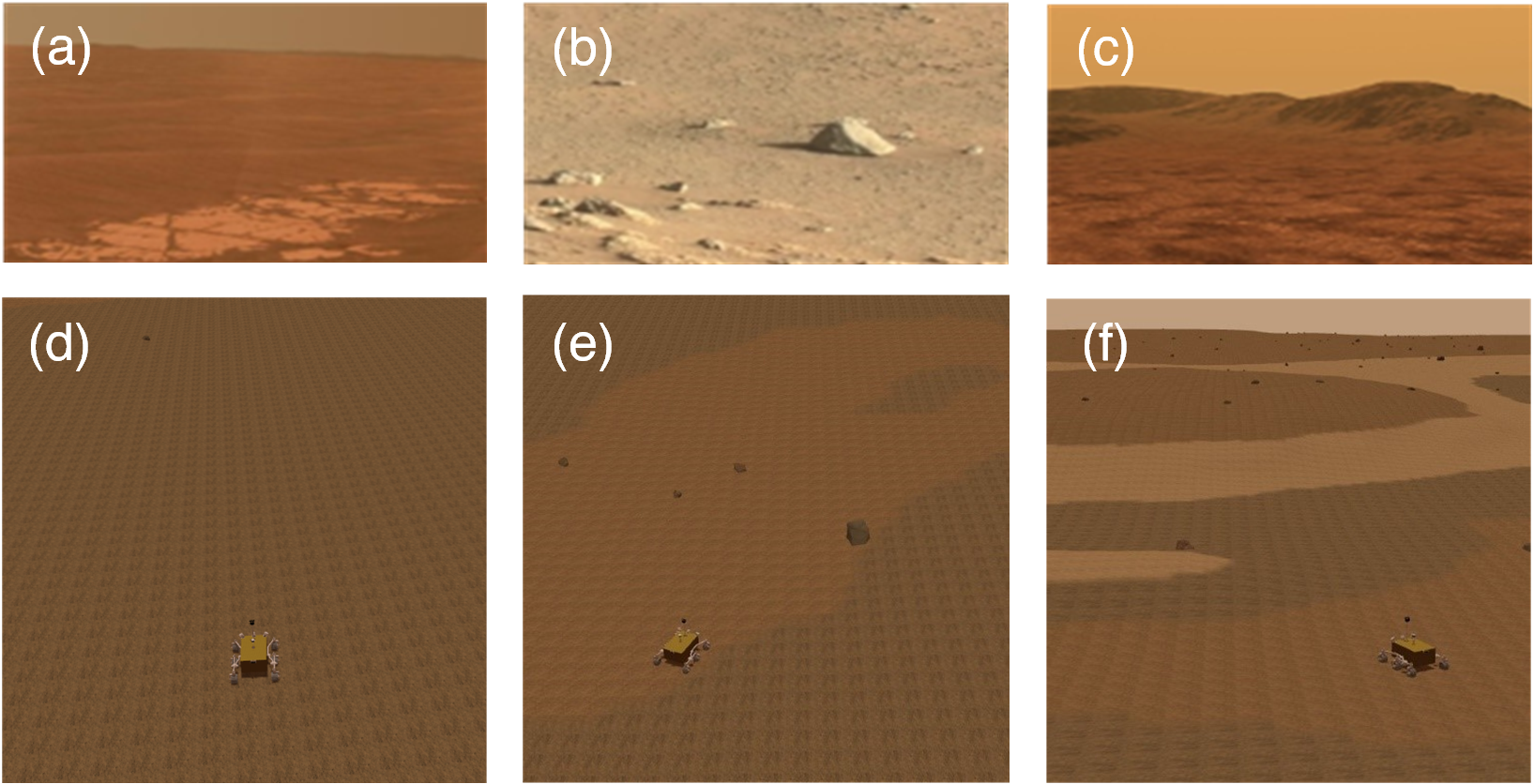}
\caption{Flat, rocky and challenging terrain as (a), (b), and (c), and the corresponding simulated environment as (d), (e), and (f).}
\label{fig_3}
\vspace {-1.5em}
\end{figure}

Three main terrain categories are designed as \figref{fig_3}: \textbf{flat} terrain with minimal elevation changes; \textbf{rocky} terrain with increased rock density for obstacles; \textbf{challenging} terrain, which includes \textit{sloping} (height variation \textless 2m with smooth slopes), \textit{rough} (high-frequency irregularities with mild elevation changes), and \textit{difficult} (maximized height variation and dense rocks) scenarios.

The terrain elevation is generated based on greyscale maps, which is created using Perlin noise\cite{perlin1985image}, a technique that allows the creation of natural looking random patterns. Parameters such as octaves and lacunarity are adjusted to control the noise complexity and frequency, affecting the elevation distribution of the terrain. Once the greyscale map is created, the grayscale values (0–255) are mapped to elevation, with a predefined height variation range. Then, rock obstacles are randomly placed based on a predefined density parameter, with varying sizes to introduce natural complexity. Finally, terrain and rock textures are applied to enhance visual quality.

\section{Experiments and Ablation Studies}

The experiments are divided into three parts: terrain classification, single-mode traversal, and multi-mode traversal. The terrain classification experiment aims to evaluate the performance of the VLM-based classifier across different terrains. In single-mode traversal tests, local navigation methods are applied to different complexity of terrain individually to assess their adaptability. Finally, the multi-mode traversal experiment utilizes the integrated multi-mode navigation system to traverse complex terrains, demonstrating the system's robustness and its ability to enhance efficiency while maintaining safety. 

\subsection{Terrain Classification}

Experiments are conducted across maps representing three levels of terrain complexity, each with four scenarios, with the rover placed at five distinct positions within each scenario. The baseline method uses a geometric classification approach based on the local elevation gridmap built with elevation mapping\cite{miki2022elevation,erni2023mem}. Rock coverage is assessed by analyzing elevation variations within a circular region and counting the grid cells where the standard deviation exceeds a given threshold. The average slope and slope variance are calculated by fitting local elevation data to planes within the same region and evaluating their inclinations. These metrics are then compared with manually set classification thresholds to classify terrain complexity. VLM classification results, along with scores of rock distribution and slope variation, are recorded as well. Using the environment complexity as ground truth, classification accuracy is evaluated for both the baseline method and the VLM approach, and overall accuracy is computed representing the performance of each method.

Results shown in Table \ref{flat classification}(a) illustrate that  flat terrain, both methods perform similarly with high accuracy, indicating minimal challenges for structured, low-complexity environments. The geometric method detects few rocks and slopes, while the VLM method also gives very low complexity scores, which is consistent with the scene characteristic.  

\begin{figure}[tbp]
\vspace {6pt}
\centering
\includegraphics[scale=1.0]{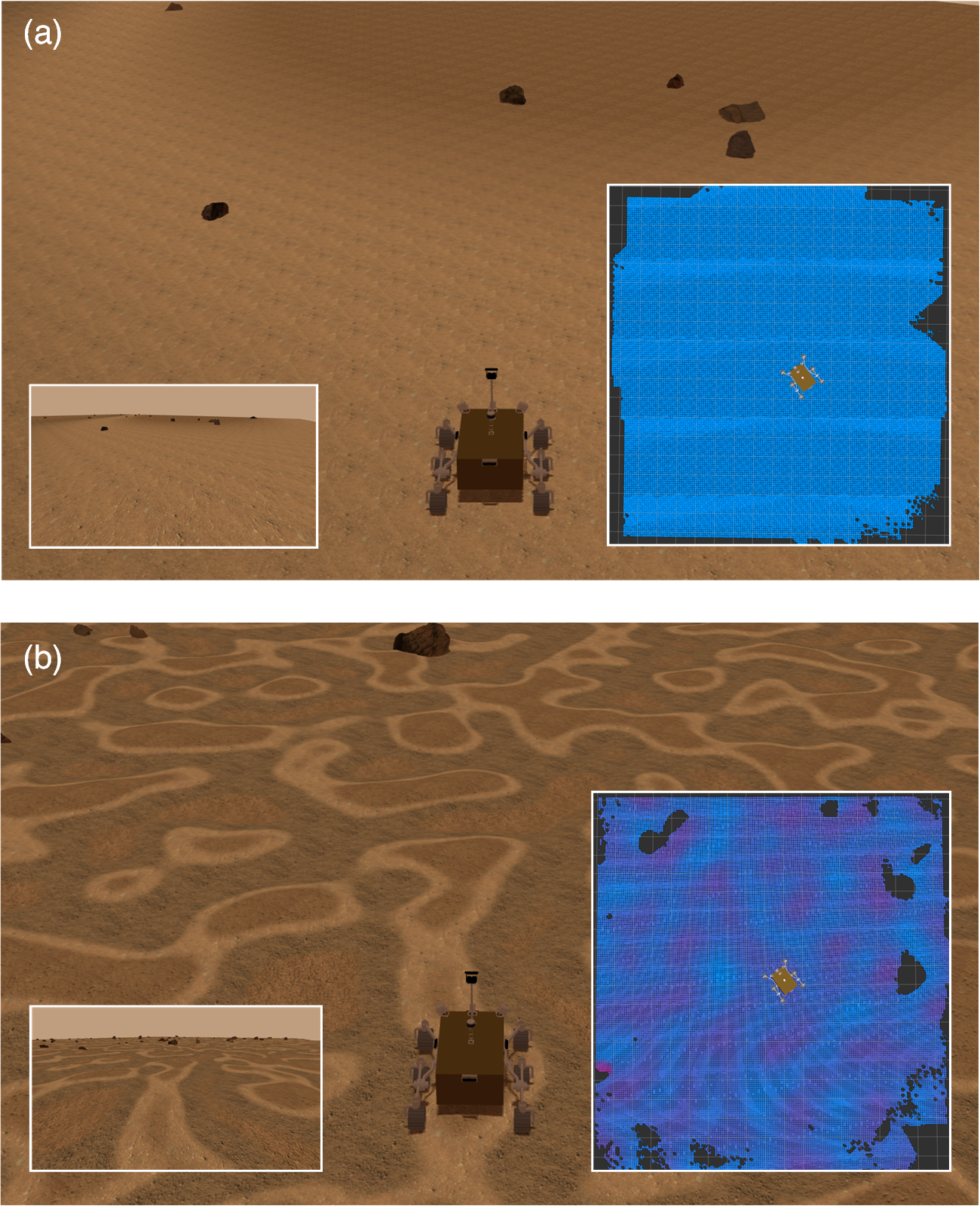}
\caption{Terrain classification environment, with RGB image for VLM input and local elevation map for geometry input. (a) rocky terrain, with RGB image and elevation map; (b) challenging terrain, with RGB image and elevation map.}
\label{fig_4}
\vspace {-1.5em}
\end{figure}

On rocky terrain, the accuracy of both geometric method and VLM method slightly decrease, but VLM remains high accuracy (above 80\%), showing its advantage in adaptability as Table \ref{flat classification}(b). In the scenario of \figref{fig_4}(a), the VLM-based approach and is able to take account of more distant rocks adequately in advance, not limited by the map construction range of the geometric-based approach, resulting in more accurate terrain complexity assessment.

On challenging terrain, the accuracy of geometric method drops significantly, especially in ambiguous scenarios with high gradient or uneven surface, shown as Table \ref{flat classification}(c). In the scenario shown as \figref{fig_4}(b), its accuracy drops to 60\%, while the VLM remains robust, achieving 100\% accuracy in all cases. This suggests that while geometric perception struggles with variation in steep and rocky landscapes, VLM better captures terrain complexity and maintains stable classification performance.

Overall, the VLM approach shows better performance in moderate and complex terrain than geometric perception. Its adaptability and stability make it a superior choice for challenging classification tasks, especially in environments with high variability in rock distribution and slope conditions.

\subsection{Single-mode Traversal}

\begin{figure*}[!ht]
\centering
\includegraphics[scale=0.95]{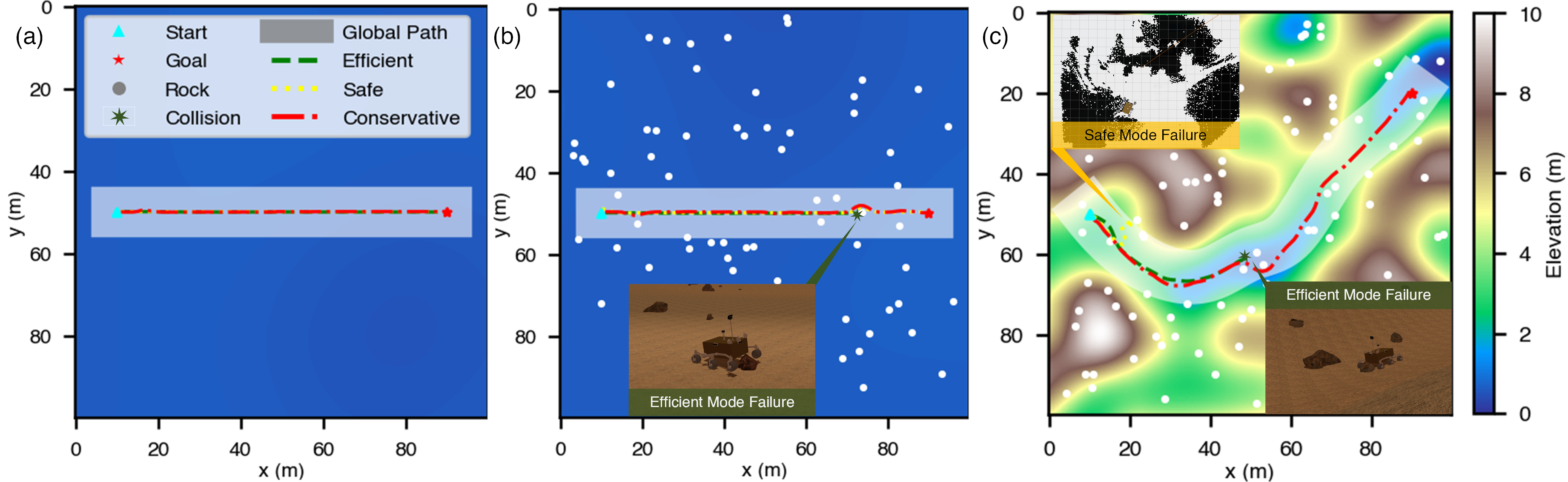}
\caption{Single-mode Traversal Results. (a) flat terrain; (b) rocky terrain, subfigure indicating that efficient mode collides with obstacles; (c) challenging terrain, subfigure indicating that efficient mode collides with obstacles while safe mode cannot find a valid path.}
\label{fig_5}
\vspace {-1.0em}
\end{figure*}

Experiments are conducted to evaluate the performance of different navigation modes across three progressively complex terrain types: flat, rocky, and challenging. Three navigation modes are used separately: efficient, safe, and conservative, shown as \figref{fig_5}.  Each mode is tested with five different traversals on each terrain, and performance is evaluated from two aspects: traversal efficiency, which measures the time taken to complete the task, and navigation safety, which records the success rate across multiple traversals, shown as Table \ref{single_mode metrics}. 

On flat terrain as \figref{fig_5}(a), all three modes successfully complete the navigation task. Due to the path optimization method and the highest tracking speed of efficient mode, its total traversal time is only 40\% of safe mode and a quarter of conservative mode, making it superior on flat terrain, providing the fastest traversal with reliability.

In contrast, on rocky terrain of \figref{fig_5}(b), the efficient mode faces significant challenges due to its lack of obstacle detection, leading to potential collisions with rocks during navigation. On the other hand, both safe mode and conservative mode are able to sense the environment and navigate safely to the destination. The safe mode adopts a higher speed between the two modes, allowing the rover to move faster and complete the task in 60\% of the time of the conservative mode.

On challenging terrain shown in \figref{fig_5}(c), the efficient mode fails to complete the task due to its inability to perceive the hazards. Safe mode's navigation algorithm also struggles with the complex landscape, misinterpreting terrain unevenness as obstacles and preventing it from planning a feasible path. On the other hand, the conservative mode maintains stable performance, successfully navigating through rough terrain and rocks for a safe arrival to the destination.

The result shows that different navigation modes perform optimally in their respective environments, providing the most efficient traversal safely. Moreover, since our multi-mode system can switch to the suitable navigation mode by adapting to the terrain, the system always operates based on the most efficient navigation method while maintaining safety, thus improving the overall navigation performance.

\begin{table*}[!htbp]
\vspace {6pt}
  \centering
  \setlength{\tabcolsep}{3pt}
  \renewcommand{\arraystretch}{1.1}
  \caption{Terrain Classification Results}
  \caption*{(a) Flat Terrain Classification}
  \begin{tabular}{c|cccc|ccc|cc|cc}
    \toprule
     \multirow{2}{*}{\centering Group} & \multicolumn{4}{c|}{Terrain Parameters} & \multicolumn{3}{c|}{Geometry-based Method} & \multicolumn{2}{c|}{VLM Method} & \multicolumn{2}{c}{Accuracy} \\
    \cmidrule(lr){2-12}
     & \multicolumn{1}{p{0.9cm}}{\centering Octaves} 
    & \multicolumn{1}{p{1.3cm}}{\centering Lacunarity} 
    & \multicolumn{1}{p{1.55cm}}{\centering Height Variation (m)} 
    & \multicolumn{1}{p{1.4cm}|}{\centering Rock Coverage} 
    & \multicolumn{1}{p{1.3cm}}{\centering Avg. Rock Grid Num.} 
    & \multicolumn{1}{p{1.4cm}}{\centering Avg. Slope Value} 
    & \multicolumn{1}{p{1.4cm}|}{\centering Avg. Slope Variance} 
    & \multicolumn{1}{p{1.5cm}}{\centering Avg. Rock Complexity} 
    & \multicolumn{1}{p{1.5cm}|}{\centering Avg. Slope Complexity} 
&Geometry &VLM \\
    \midrule
    1  & 3.0& 1.0& 0.2  & 0.01  & 0  & 0.308 & 0.214 & 0.04  & 0.04  &100\% & 100\% \\
    2  & 3.0& 1.0& 0.2  & 0.01  & 0  & 0.346 & 0.2   & 0     & 0.08  &100\% & 100\% \\
    3  & 3.0& 1.0& 1.0& 0.01  & 0  & 3.148 & 2.466 & 0.04  & 0.1   &100\% & 100\% \\
    4  & 3.0& 1.0& 2.0& 0.02  & 0  & 6.924 & 15.8  & 0.1   & 0.16  &100\% & 100\% \\
    \bottomrule
  \end{tabular}%
  \label{flat classification}
\end{table*}%

\begin{table*}[!htbp]
  \centering
  \setlength{\tabcolsep}{3pt} 
  \renewcommand{\arraystretch}{1.1} 
  \caption*{(b) Rocky Terrain Classification}
  \begin{tabular}{c|cccc|ccc|cc|cc}
    \toprule
     \multirow{2}{*}{\centering Group} & \multicolumn{4}{c|}{Terrain Parameters} & \multicolumn{3}{c|}{Geometry-based Method} & \multicolumn{2}{c|}{VLM Method} & \multicolumn{2}{c}{Accuracy} \\
    \cmidrule(lr){2-12}
     & \multicolumn{1}{p{0.9cm}}{\centering Octaves} 
    & \multicolumn{1}{p{1.3cm}}{\centering Lacunarity} 
    & \multicolumn{1}{p{1.55cm}}{\centering Height Variation (m)} 
    & \multicolumn{1}{p{1.4cm}|}{\centering Rock Coverage} 
    & \multicolumn{1}{p{1.3cm}}{\centering Avg. Rock Grid Num.} 
    & \multicolumn{1}{p{1.4cm}}{\centering Avg. Slope Value} 
    & \multicolumn{1}{p{1.4cm}|}{\centering Avg. Slope Variance} 
    & \multicolumn{1}{p{1.5cm}}{\centering Avg. Rock Complexity} 
    & \multicolumn{1}{p{1.5cm}|}{\centering Avg. Slope Complexity} 
&Geometry &VLM \\
    \midrule
    1   & 3.0& 1.0& 0.2   & 0.04  & 125   & 4.152 & 52.512 & 0.32  & 0.1   & 60\%  & \textbf{80\%} \\
    2   & 3.0& 1.0& 0.2   & 0.06  & 393.8 & 3.714 & 138.624 & 0.66  & 0.18  & 100\% & 100\% \\
    3   & 3.0& 1.0& 1.0& 0.06  & 401   & 6.394 & 138.244 & 0.66  & 0.3   & 100\% & 100\% \\
    4   & 3.0& 1.0& 1.0& 0.08  & 698.8 & 7.778 & 236.166 & 0.72  & 0.22  & 100\% & 100\% \\
    \bottomrule
  \end{tabular}%
  \label{rocky classification}
\end{table*}%

\begin{table*}[htbp]
  \centering
  \setlength{\tabcolsep}{3pt} 
  \renewcommand{\arraystretch}{1.1} 
  \caption*{(c) Challenging Terrain Classification}
  \begin{tabular}{c|cccc|ccc|cc|cc}
    \toprule
     \multirow{2}{*}{\centering Group} & \multicolumn{4}{c|}{Terrain Parameters} & \multicolumn{3}{c|}{Geometry-based Method} & \multicolumn{2}{c|}{VLM Method} & \multicolumn{2}{c}{Accuracy} \\
    \cmidrule(lr){2-12}
     & \multicolumn{1}{p{0.9cm}}{\centering Octaves} 
    & \multicolumn{1}{p{1.3cm}}{\centering Lacunarity} 
    & \multicolumn{1}{p{1.55cm}}{\centering Height Variation (m)} 
    & \multicolumn{1}{p{1.4cm}|}{\centering Rock Coverage} 
    & \multicolumn{1}{p{1.3cm}}{\centering Avg. Rock Grid Num.} 
    & \multicolumn{1}{p{1.4cm}}{\centering Avg. Slope Value} 
    & \multicolumn{1}{p{1.4cm}|}{\centering Avg. Slope Variance} 
    & \multicolumn{1}{p{1.5cm}}{\centering Avg. Rock Complexity} 
    & \multicolumn{1}{p{1.5cm}|}{\centering Avg. Slope Complexity} 
&Geometry &VLM \\
    \midrule
    1   & 3.0& 1.0& 10.0& 0.02  & 835.4 & 31.702 & 276.424 & 0.18  & 0.74  & 80\%  & \textbf{100\%} \\
    2   & 10.0& 1.5   & 1.0& 0.06  & 344.6 & 20.514 & 235.796 & 0.74  & 0.68  & 60\%  & \textbf{100\%} \\
    3   & 3.0& 2.0& 5.0& 0.05  & 414.2 & 29.314 & 313.148 & 0.4   & 0.7   & 100\% & 100\% \\
    4   & 3.0& 2.0& 5.0& 0.04  & 183.2 & 38.942 & 268.054 & 0.42  & 0.66  & 100\% & 100\% \\
    \bottomrule
  \end{tabular}%
  \label{challenging classification}
\label{terrain classification metrics}
\end{table*}%

\subsection{Multi-mode Traversal}

In order to evaluate the adaptability of our multi-mode navigation system across diverse terrain, we design a large-scale mixed terrain scenario which consists of three terrain types. For benchmarking, we set a baseline approach based on the conservative mode, and compare it with our multi-mode system, which autonomously switches between modes with a VLM-driven strategy. During navigation, the time and distance under different modes are recorded, along with total time and distance, shown as Table \ref{multi_mode metrics}.

\begin{figure}[!t]
\centering
\includegraphics[scale=1.0]{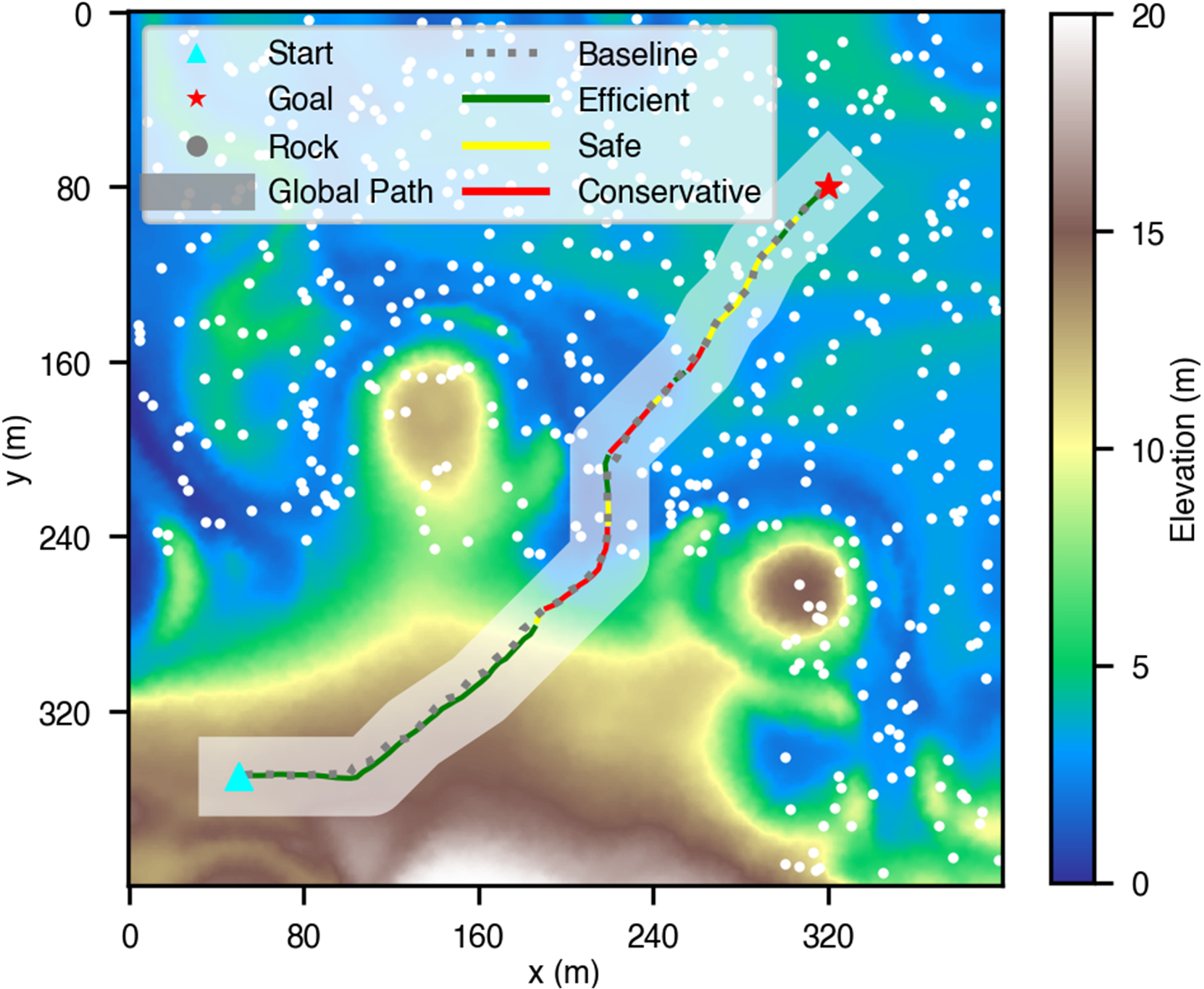}
\caption{Multi-mode Traversal Result.}
\label{fig_6}
\vspace {-2.0em}
\end{figure}

As the result in \figref{fig_6} shows, both methods successfully traverse the complex terrain. For the multi-mode system, during the entire traversal, efficient mode, safe mode and conservative mode accounted for 52.2\%, 23.0\% and 24.8\% of the total distance respectively, while consuming 24\%, 26.3\% and 49.7\% of the total time. Due to the multi-mode navigation system's ability to dynamically switch between different modes based on terrain conditions, it achieves significantly shorter traversal times at a similar distance, consuming only 55.7\% of the time required by the baseline mode, with efficiency improvement as 79.5\%, assuming the efficiency improvement of multi-mode navigation while maintaining safe traversal. These results show that the multi-mode navigation system excels in complex terrain, improving navigation efficiency without compromising safety.

\section{Conclusion}
This paper presents a VLM-empowered multi-mode system for efficient and safe planetary navigation. In contrast to traditional geometric perception methods, the proposed approach uses Vision-Language Model for terrain complexity classification and demonstrates superior performance in complex environments. The method also offers enhanced flexibility, allowing task-specific and vehicle-specific adaptation. Three different navigation strategies corresponding to different terrain types are developed and an integrated system is constructed for seamless management. Finally, simulation experiments on single-mode and multi-mode traversals demonstrate that the multi-mode system dynamically adjusts the navigation mode based on terrain features, enhancing efficiency while maintaining safety, achieving a 79.5\% improvement in efficiency compared to the single-mode conservative navigation system.

While the system exhibits considerable extensibility, the current algorithmic implementation focuses primarily on the geometric properties of the terrain, with limited attention paid to semantic aspects. Future work will aim to refine the algorithms for each mode and proceed with real-world deployment to assess the robustness of the system in different real-world environments.

\begin{table}[!t]
\vspace {6pt}
  \centering
  \setlength{\tabcolsep}{3pt} 
  \renewcommand{\arraystretch}{1.1} 
  \caption{Single-mode Traversal Metrics}
  \begin{tabular}{c|>{\centering\arraybackslash}p{2cm} >{\centering\arraybackslash}p{1.15cm} >{\centering\arraybackslash}p{1.15cm} >{\centering\arraybackslash}p{1.45cm} }
    \toprule
      
    &  metrics
    & Efficient& Safe& Conservative\\ 

    \midrule
    \multirow{2}{*}{Scene a} & Success Rate & \textbf{100}\% & 100\% & 100\% \\
    & Avg.  Time (s) & \textbf{42.5} & 110.8 & 178.5 \\
     \midrule
    \multirow{2}{*}{Scene b} & Success Rate & 20\% & \textbf{100}\% & 100\% \\
    & Avg. Time (s) & - & \textbf{116.5} & 189 \\
    \midrule
    \multirow{2}{*}{Scene c} & Success Rate & 0\% & 0\% & \textbf{100}\% \\
    & Avg.  Time(s)& - & - & \textbf{293} \\
    \bottomrule
  \end{tabular}
  \label{single_mode metrics}
\end{table}

\begin{table}[!t]
  \centering
  \setlength{\tabcolsep}{3pt} 
  \renewcommand{\arraystretch}{1.1} 
  \caption{Multi-mode Traversal Metrics.}
  \begin{tabular}{c|c|c|ccc}
    \toprule
    \multirow{2}{*}{Metrics} & \multirow{2}{*}{\centering Single-mode}& \multicolumn{4}{c}{Multi-mode} \\
    \cmidrule(lr){3-6}
     & & Total &Efficient& Safe& Conservative\\
    \midrule
    Time(s) & \textbf{1081.7} & \textbf{602.6} & 144.9 & 158.6 & 299.1 \\
    Distance(m)& \textbf{413.7}&  \textbf{411.8}& 215.1 & 94.7 & 102.0  \\
    \bottomrule
  \end{tabular}%
  \label{multi_mode metrics}%
\end{table}

\bibliographystyle{unsrt}
\bibliography{references}

\end{document}